\documentclass[sigconf]{acmart}

\usepackage{subcaption}
\usepackage{multirow}
\usepackage[linesnumbered,ruled]{algorithm2e}
\AtEndPreamble{
    \usepackage[capitalize]{cleveref}
    \crefname{section}{Sec.}{Secs.}
    \Crefname{section}{Section}{Sections}
    \crefname{table}{Tab.}{Tabs.}
    \Crefname{table}{Table}{Tables}
}

\usepackage{xspace}
\newcommand{\etal}{\textit{et al.}\xspace}
\newcommand{\eg}{\textit{e.g.}\xspace}
\newcommand{\ie}{\textit{i.e.}\xspace}

\newcommand{\etc}{\textit{etc.}\xspace}
\usepackage{enumitem}

\AtBeginDocument{%
  }

\setcopyright{rightsretained}
\copyrightyear{2024}
\acmYear{2024}
\acmDOI{10.1145/3664647.3685510}

\acmConference[MM '24]{Proceedings of the 32nd ACM International Conference on Multimedia}{October 28-November 1, 2024}{Melbourne, VIC, Australia}

\acmBooktitle{Proceedings of the 32nd ACM International Conference on Multimedia (MM '24), October 28-November 1, 2024, Melbourne, VIC, Australia}
\acmISBN{979-8-4007-0686-8/24/10}

\makeatletter
\gdef\@copyrightpermission{
  \begin{minipage}{0.3\columnwidth}
   \href{https://creativecommons.org/licenses/by/4.0/}{\includegraphics[width=0.90\textwidth]{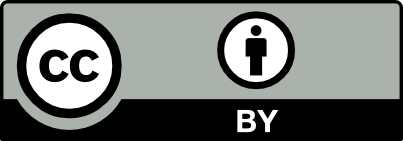}}
  \end{minipage}\hfill
  \begin{minipage}{0.7\columnwidth}
   \href{https://creativecommons.org/licenses/by/4.0/}{This work is licensed under a Creative Commons Attribution International 4.0 License.}
  \end{minipage}
  \vspace{5pt}
}
\makeatother

\begin{document}

\title[PIP: Detecting Adversarial Examples in LVLMs via Attention Patterns of Irrelevant Probe Questions]{PIP: Detecting Adversarial Examples in Large Vision-Language Models via Attention Patterns of Irrelevant Probe Questions}

\author{Yudong Zhang}
\orcid{0009-0009-6049-603X}

\affiliation{%
  % \institution{Electrical Engineering Department of Tsinghua University}
  \institution{Tsinghua University}
  \institution{\& Tencent}
  \city{Beijing}
  \country{China}
}
\email{zhangyd16@mails.tsinghua.edu.cn}

\author{Ruobing Xie}
\orcid{0000-0003-3170-5647}
\authornote{Corresponding author.}
\affiliation{%
  \institution{Tencent}
  \city{Beijing}
  \country{China}}
\email{xrbsnowing@163.com}

\author{Jiansheng Chen}
\orcid{0000-0002-2040-7938}
\authornotemark[1]
\affiliation{%
  \institution{University of Science and Technology Beijing}
  \city{Beijing}
  \country{China}
}
\email{jschen@ustb.edu.cn}

\author{Xingwu Sun}
\orcid{0009-0008-3222-0901}
\affiliation{%
 \institution{University of Macau}
 \institution{\& Tencent}
 \city{Beijing}
 \country{China}}
\email{sunxingwu01@gmail.com}

\author{Yu Wang}
\orcid{0000-0001-6108-5157}
\authornotemark[1]
\affiliation{%
  \institution{Tsinghua University}
  \city{Beijing}
  \country{China}}
\email{yu-wang@mail.tsinghua.edu.cn}

% \renewcommand{\shortauthors}{Zhang et al.}

\begin{abstract}
  Large Vision-Language Models (LVLMs) have demonstrated their powerful multimodal capabilities. However, they also face serious safety problems, as adversaries can induce robustness issues in LVLMs through the use of well-designed adversarial examples. Therefore, LVLMs are in urgent need of detection tools for adversarial examples to prevent incorrect responses. In this work, we first discover that LVLMs exhibit regular attention patterns for clean images when presented with probe questions. We propose an unconventional method named PIP, which utilizes the attention patterns of one randomly selected irrelevant probe question (e.g., ``Is there a clock?'') to distinguish adversarial examples from clean examples. Regardless of the image to be tested and its corresponding question, PIP only needs to perform one additional inference of the image to be tested and the probe question, and then achieves successful detection of adversarial examples. Even under black-box attacks and open dataset scenarios, our PIP, coupled with a simple SVM, still achieves more than 98\% recall and a precision of over 90\%. Our PIP is the first attempt to detect adversarial attacks on LVLMs via simple irrelevant probe questions, shedding light on deeper understanding and introspection within LVLMs. The code is available at \url{https://github.com/btzyd/pip}.
\end{abstract}

\begin{CCSXML}
<ccs2012>
   <concept>
       <concept_id>10002978.10002997.10002999</concept_id>
       <concept_desc>Security and privacy~Intrusion detection systems</concept_desc>
       <concept_significance>300</concept_significance>
       </concept>
   <concept>
       <concept_id>10010147.10010178.10010224</concept_id>
       <concept_desc>Computing methodologies~Computer vision</concept_desc>
       <concept_significance>500</concept_significance>
       </concept>
   <concept>
       <concept_id>10010147.10010178.10010179</concept_id>
       <concept_desc>Computing methodologies~Natural language processing</concept_desc>
       <concept_significance>100</concept_significance>
       </concept>
 </ccs2012>
\end{CCSXML}

\ccsdesc[300]{Security and privacy~Intrusion detection systems}

\keywords{Large Vision-Language Model, Detecting Adversarial Example.}


\maketitle

\section{Introduction}
Large vision-language models (LVLMs) have demonstrated their powerful multimodal capabilities across a range of tasks \cite{liu2024visual, wang2024visionllm, alayrac2022flamingo}. However, LVLMs continue to confront significant security challenges. Adversaries can perturb the images with elaborate and almost imperceptible noise, leading LVLMs to produce incorrect outputs or even outputs that align with the adversaries' intentions \cite{carlini2024aligned, bailey2023image, qi2024visual, zhao2024evaluating, yin2023vlattack, dong2023robust}. Significant security issues have impeded the adoption of LVLMs in critical areas \cite{10.1145/3581783.3612337, 10.1145/3581783.3611914, 10.1145/3503161.3547801, 10.1145/3626772.3657781, luo2023image, lu2023set, 10.1145/3503161.3547989}.

\begin{figure}[t]
    \centering
    \includegraphics[width=\linewidth]{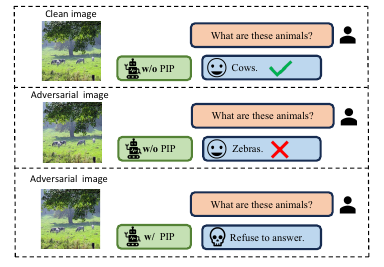}
    \caption{Implications for our adversarial example detection method PIP. (\emph{Top}): LVLMs can give correct answers for clean images. (\emph{Middle}): LVLMs may give incorrect answers for adversarial images. (\emph{Down}): When detecting adversarial examples through our simple PIP, LVLMs reject answers for adversarial examples to prevent security risks.}
    \label{fig:intro}
\end{figure}

Adversarial attacks on image modalities are facilitated by their high-dimensional and continuous nature. In recent years, there has been a surge in studies on adversarial attacks on LVLMs. Attack-Bard \cite{dong2023robust} manipulates images to make ChatGPT-4 errors on image caption, while Carlini \cite{carlini2024aligned} leads LVLMs to produce unethical words in response to adversarial images.
In contrast to adversarial attacks, there has been limited research on the detection of adversarial examples. Therefore, there is an urgent need for detection methods for adversarial examples to enhance the safety of LVLMs.

\begin{figure*}[t]
    \centering
    \includegraphics[width=0.96\textwidth]{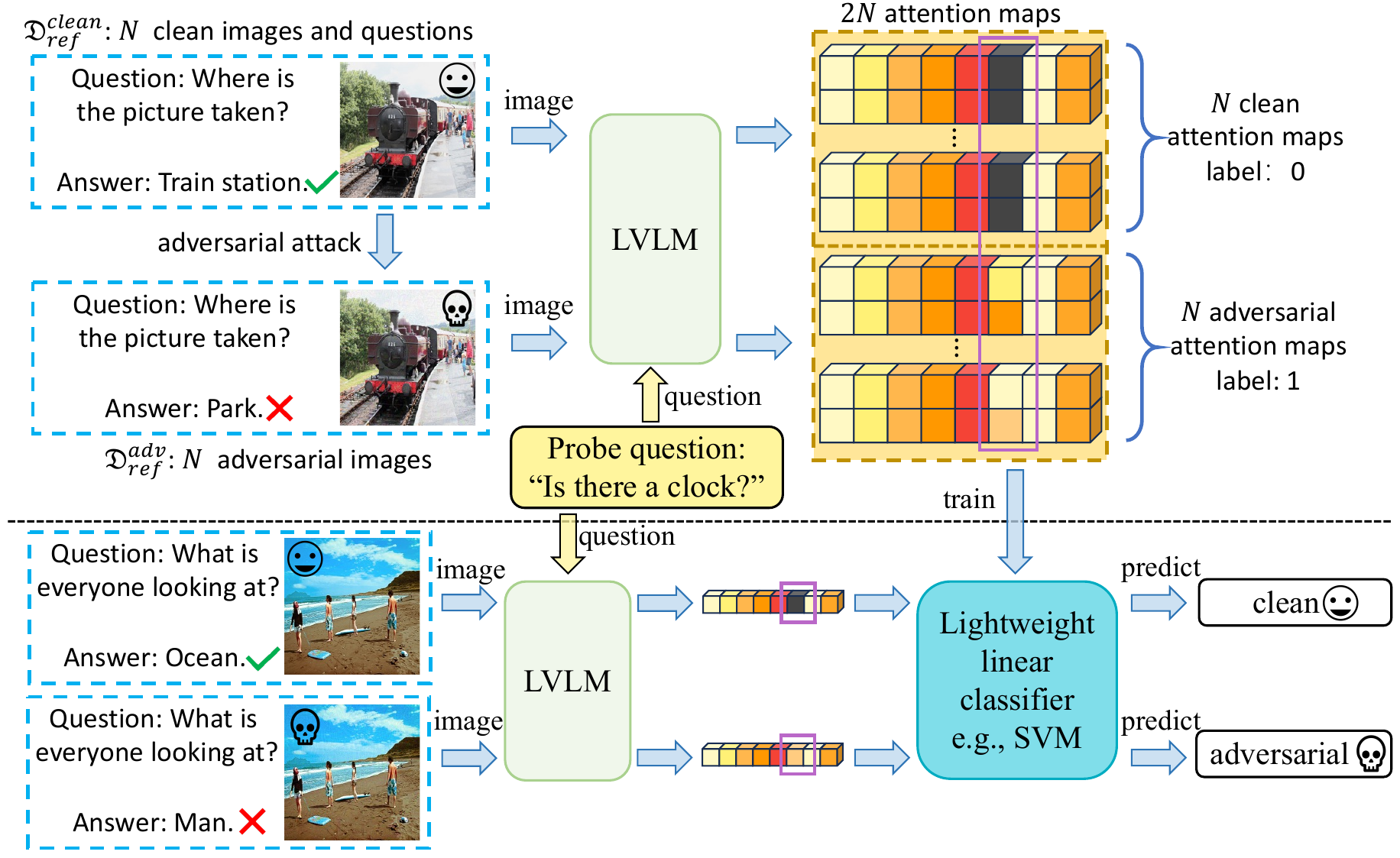}
    \caption{The pipeline of our proposed PIP. The top half is operated offline, while the bottom half is operated online. (\emph{Top}): We perform adversarial attacks on $N$ images in $\mathcal{D}^{clean}_{ref}$ and obtain $N$ adversarial images, which constitute $\mathcal{D}^{adv}_{ref}$. We extract their attention maps (attention of the first word generated by LLM to all image tokens) of the LVLM with the irrelevant probe question ``Is there a clock'', and train a lightweight linear classifier (\eg, SVM) with these $2N$ attention maps. (\emph{Down}): For images to be tested from $\mathcal{D}_{test}$, we first get their attention maps with the same probe question, and use the classifier to determine whether they are adversarial examples or not. Surprisingly, this simple method PIP functions well in this challenging task.}
    \label{fig:overall}
\end{figure*}

Previous work \cite{pang2018towards, xu2017feature, meng2017magnet} has primarily focused on the detection of adversarial examples in isolated vision models, \ie, image classifiers based on convolutional neural networks (CNNs), which are not applicable to LVLMs. The detection of adversarial examples on LVLMs faces several challenges: First, LVLMs have more layers and contain interactions between image and text messages, making it difficult to determine where adversarial attacks take effect. Second, traditional CNNs have a more centralized data distribution (\eg, imagenet classifiers detect imagenet adversarial samples), while LVLMs accept data inputs from open scenarios. In addition, LVLMs have a high inference cost and it is inappropriate to introduce too many inference times during the detection process. Few studies have focused on the detection of adversarial examples on LVLMs.

In this paper, we introduce an extremely simple method for the detection of adversarial examples, named \textbf{PIP}, that leverages the \textbf{attention \underline{p}attern of \underline{i}rrelevant \underline{p}robe questions}. We initially discovered that for yes/no type questions, LVLM exhibits a \emph{regular attention pattern} for image tokens when generating text output. Subsequently, we observed markedly distinct, linearly separable attention patterns between clean and adversarial images in response to the same simple yes/no probe questions. Building on this observation, we trained linear classifiers to distinguish between clean and adversarial examples, utilizing their attention patterns on one simple irrelevant probe question. Our PIP does not require the determination of whether an image should respond with ``yes'' or ``no'' to irrelevant probe questions, as our focus is exclusively on analyzing the attention patterns of the image in response to these questions, rather than the answers.

The workflow of PIP is illustrated in \cref{fig:overall}. We proposed our very simple and straightforward PIP detector could detect the sophisticated well-designed diverse adversarial examples of LVLMs. Our approach does not need to consider the distribution of the images, the type of the questions, the adversarial attack method, \etc, but only needs to ask LVLMs one probe question on examples, and trains an SVM \cite{boser1992training} in the offline phase by using the attention patterns of clean and adversarial examples on the probe question. Surprisingly, it achieves the detection of adversarial examples for a wide range of questions, even if the probe question is randomly chosen and irrelevant. For the images entered by users in the online phase, the attention maps of images and the probe question are obtained by one additional inference, and the clean and adversarial examples can be distinguished via the lightweight SVM classifier.

Our main contributions can be summarized as follows:
\begin{itemize}[leftmargin=*]
    \item We have defined a new task: adversarial example detection for image adversarial attacks on LVLMs, which is essential in practice. We have also shown how to adapt the basic multimedia task evaluation metrics to this task.
    \item To the best of our knowledge, we are the first to introduce the simple and unconventional PIP, which detects adversarial examples based on the attention pattern of irrelevant probe questions. Extensive experiments in different settings have validated the effectiveness, universality, and transferability of our PIP. 
    \item PIP can also inspire subsequent work, such as defense and purification against samples. It can also help us understand LVLM in terms of deeper mechanisms.
\end{itemize}

\section{Related Works}

\subsection{Large Vision-Language Models}
Alignment-based vision-language models generally comprise three modules: a visual encoder, a large language model, and a vision-language alignment module. Generally, LVLMs utilize pre-trained visual encoders and large language models, with the vision-language alignment module fine-tuned to enable multimodal capabilities. Recently, a lightweight vision-language alignment module, the Querying Transformer (Q-former), has gained popularity. Recent popular LVLMs employing this alignment technique include BLIP-2 \cite{li2023blip}, InstructBLIP \cite{dai2024instructblip}, and MiniGPT-4 \cite{zhu2023minigpt}.

\subsection{Adversarial Attacks and Adversarial Examples}
Adversarial attacks generate adversarial examples by introducing almost imperceptible perturbations to images, leading neural networks to respond incorrectly. Previous studies have focused on perturbations of visual modalities, including FGSM \cite{goodfellow2014explaining}, PGD \cite{madry2017towards}, JSMA \cite{papernot2016limitations}, and C\&W attack \cite{carlini2017towards}, among others. Research has also been conducted on attacks targeting textual modalities, such as Bert-Attack \cite{li2020bert} and TextFooler \cite{jin2019bert}. Large vision-language models are also vulnerable to adversarial examples. For instance, Attack-Bard \cite{dong2023robust} induces incorrect captions in Google's Bard and OpenAI's ChatGPT-4 through black-box attacks on the image, while Carlini \etal \cite{carlini2024aligned} prompt LVLMs to generate inappropriate speech (\eg, profanity, biased statements, \etc) via white-box attacks on the image.

\subsection{Detecting Adversarial Examples}
Since the proposal of adversarial attacks, significant research has been devoted to the detection of adversarial examples, aiming to alert the model to the presence of such examples. This is particularly critical in LVLMs because deliberately designed adversarial examples can cause LLMs to generate outputs aligned with the adversary's intentions, potentially leading to robustness and hallucination issues. More gravely, the model may produce statements that are socially and morally reprehensible.

Previous research has concentrated on detecting adversarial examples for visual modality CNNs, as evidenced by \cite{dathathri2018detecting, li2017adversarial, gao2021maximum, ma2019metaadvdet}. However, research on detecting adversarial examples in vision-language multimodal models is lacking. This represents a considerable risk when employing LVLMs in sensitive domains. LVLMs require the capability to detect adversarial examples and, in response, should either refuse to answer or neutralize the adversarial input to provide a clean response.

\section{A New Task: Detecting Adversarial Examples in Large Vision-Language Models}

\subsection{Definition of Our Adversarial Examples Detection Task}
\label{sec:task_definition}

In this section, we introduce a novel task aimed at detecting adversarial examples in LVLMs.

Let $f$ represent a large vision-language model, which takes an image $I$ and a question $Q$ as input, and produces an answer $A=f(I, Q)$. The task will provide two datasets, $\mathcal{D}_{ref}$ and $\mathcal{D}_{test}$, where $\mathcal{D}_{ref}$ comprises $N$ samples from a large clean dataset $\mathcal{D}$, serving as a reference for clean examples. Additionally, we randomly select $M$ images and questions from $\mathcal{D}$ to create the dataset $\mathcal{D}_{clean}$, and then execute adversarial attacks on these $M$ images to generate the dataset $\mathcal{D}_{adv}$. By randomly combining clean and adversarial examples in a ratio of $M_{clean}:M_{adv}$, we generate the test dataset $\mathcal{D}_{test}$. The task's objective is to train a classifier $h(I)$ that discerns whether each $I\in\mathcal{D}_{test}$ originates from the clean dataset $\mathcal{D}_{clean}$ or the adversarial dataset $\mathcal{D}_{adv}$. Note that for this task, only $\mathcal{D}_{ref}$, $f$, and the test dataset $\mathcal{D}_{test}$ are provided. If $I$ originates from $\mathcal{D}_{clean}$, the ground truth $g(I)$ should be $1$, otherwise $0$, as delineated in \cref{eq:gi}.
\begin{equation}
    g(I) = 
    \begin{cases} 
      1 & \text{if } I\in\mathcal{D}_{adv}, \\
      0 & \text{if } I\in\mathcal{D}_{clean}.
    \end{cases}
    \label{eq:gi}
\end{equation}

The adversarial attacks on LVLMs are more diverse in terms of attack methods and targets, and more challenging to detect.

\subsection{Evaluation of Our Detection Task}

To evaluate the performance of the classifier $h$, which determines the likelihood of $I$ being an adversarial example, we employ the metrics outlined in \cref{eq:evaluation}.
\begin{equation}
    \begin{aligned}
        TP &= \sum_{I\in\mathcal{D}_{test}}{g(I)\cdot h(I)}, &FN = M_{adv} - TP\\
        FP &= \sum_{I\in\mathcal{D}_{test}}{(1-g(I))\cdot h(I)}, &TN = M_{clean} - FP \\
    \end{aligned}
    \label{eq:evaluation}
\end{equation}
\begin{figure}[t]
    \centering
    \begin{subfigure}{\linewidth}
        \includegraphics[width=\linewidth]{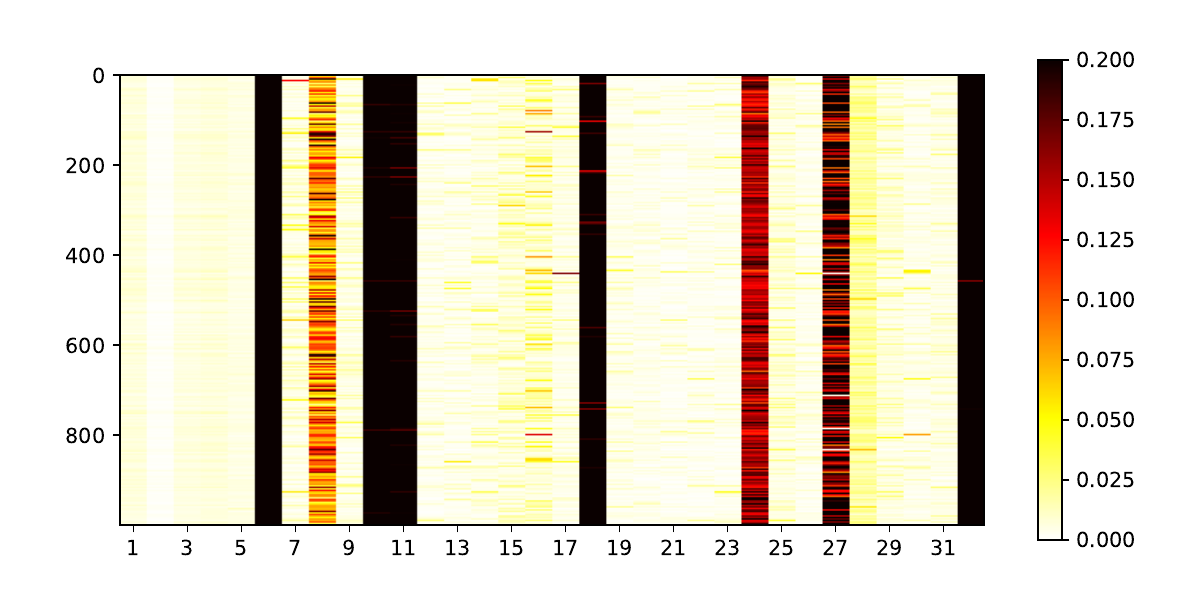}
        \caption{The attention map of ``yes/no'' questions.}
        \label{fig:attention_yesno}
    \end{subfigure}
    \hfill
    \begin{subfigure}{\linewidth}
        \includegraphics[width=\linewidth]{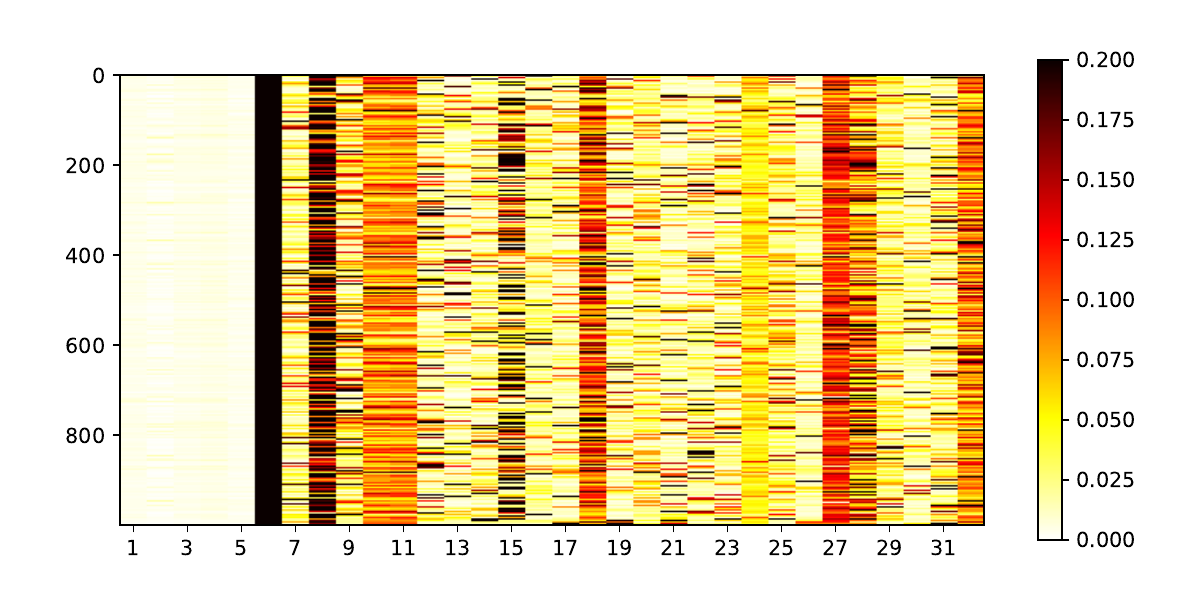}
        \caption{The attention map of ``number'' questions.}
        \label{fig:attention_number}
    \end{subfigure}
    \hfill
    \begin{subfigure}{\linewidth}
        \includegraphics[width=\linewidth]{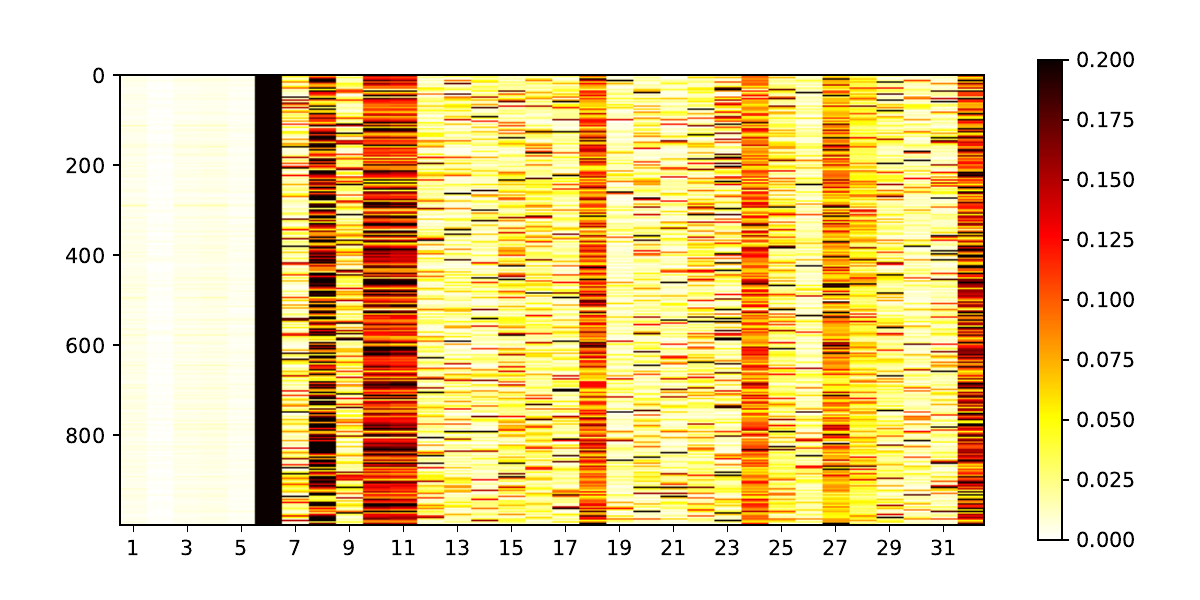}
        \caption{The attention map of ``other'' questions.}
        \label{fig:attention_other}
    \end{subfigure}
    \caption{The attention maps of different types of questions on 1,000 randomly-selected images and questions. Due to space limitations, we select only one layer (the 16th layer of the LLM) and display the maximum value in the multi-head attention. The attention map of (a) ``yes/no'' is more regular than (b) ``number'' and (c) ``other'', indicating that the simple ``yes/no'' is a more suitable probe question.}
    \label{fig:attention_diff_ques_type}
\end{figure}

Upon defining True Positives (TP), True Negatives (TN), False Positives (FP), and False Negatives (FN), we compute the precision, recall, accuracy, and F1-score to comprehensively assess the performance of classifier $h$.

\section{Explore the Use of Our PIP to Detecting Adversarial Examples}

\subsection{LVLMs Have Regularized Attention Patterns of Clean Examples to Yes/No Questions}
\label{sec:yesno_regular}

We begin by selecting a popular LVLM InstructBLIP Vicuna-7B (decoder-only) to explore attention patterns, then observe other LVLMs if there is a similar phenomenon. It is based on the Q-former architecture, and the LLM receives 32 image tokens and $P$ question tokens as inputs. Initially, the LLM computes self-attention on the $32+P$ tokens, followed by the insertion of the start token ${eos}_{start}$ into the LLM to generate the output sequence sequentially. We examine the attention map from the first token following the start token to the 32 image tokens. The attention map has a dimension of $[32, 32, 32]$, with the first $32$ representing the number of layers, the second $32$ indicating the number of multi-heads, and the last $32$ denoting the index of image tokens. To display the attention map of multiple images within a single figure, we select a certain layer and focus on the largest head within the multi-head attention.

We randomly selected 1000 questions of each type (yes/no, number, and other) from VQA v2. For each of these question types, we generated the attention maps for layer 16 of the LLM in InstructBLIP Vicuna-7B, as illustrated in \cref{fig:attention_diff_ques_type}, where each row is an image and each column is the 32 image tokens fed into the LLM, with the color indicating attention values. The attention map for yes/no questions exhibits a more regular pattern compared to those for number and other question types. This insight offers a novel perspective on distinguishing between clean and adversarial examples through the analysis of regular attention patterns.

\subsection{The Attention Patterns of ``yes/no'' Probe Questions between Clean and Adversarial Examples are Clearly Distinguishable}
\label{sec:yesno_between_clean_adv}
\begin{figure}[t]
    \centering
    \begin{subfigure}{\linewidth}
        \includegraphics[width=\linewidth]{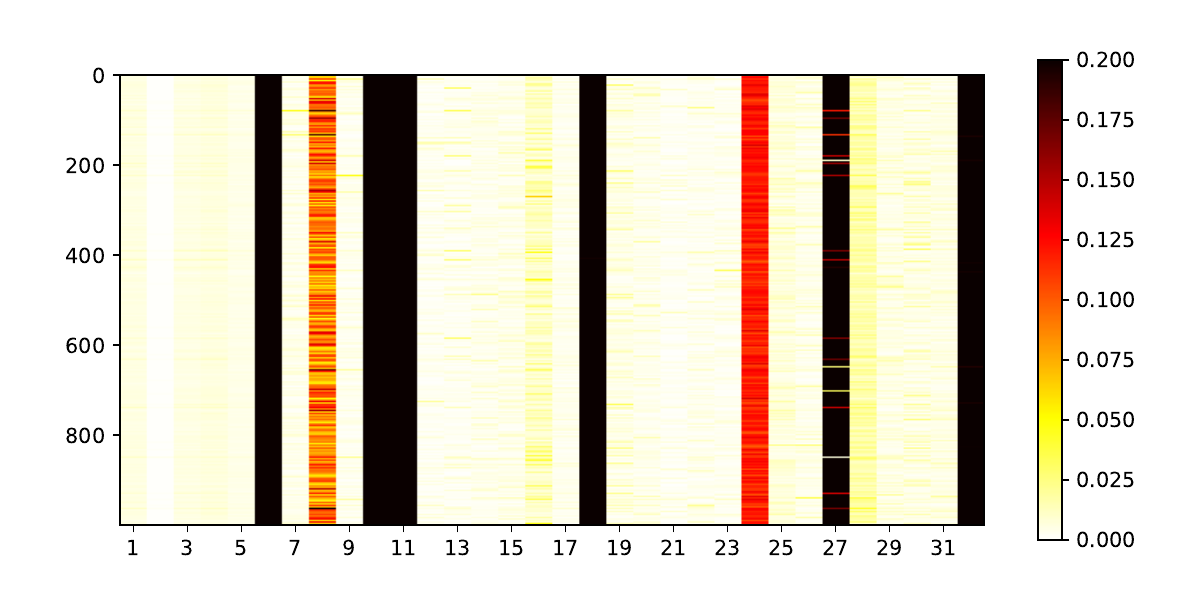}
        \caption{The attention map of $\mathcal{D}^\text{1k}_\text{clean}$.}
        \label{fig:attention_clean}
    \end{subfigure}
    \hfill
    \begin{subfigure}{\linewidth}
        \includegraphics[width=\linewidth]{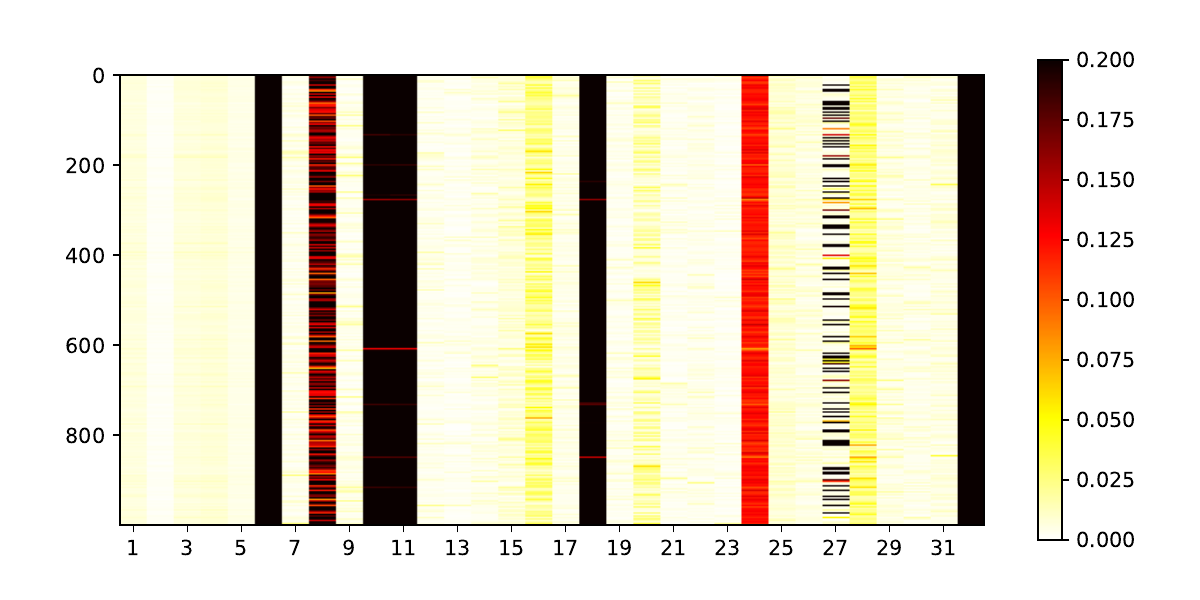}
        \caption{The attention map of $\mathcal{D}^{1k}_\text{CLIP}$.}
        \label{fig:attention_vout}
    \end{subfigure}
    \hfill
    \begin{subfigure}{\linewidth}
        \includegraphics[width=\linewidth]{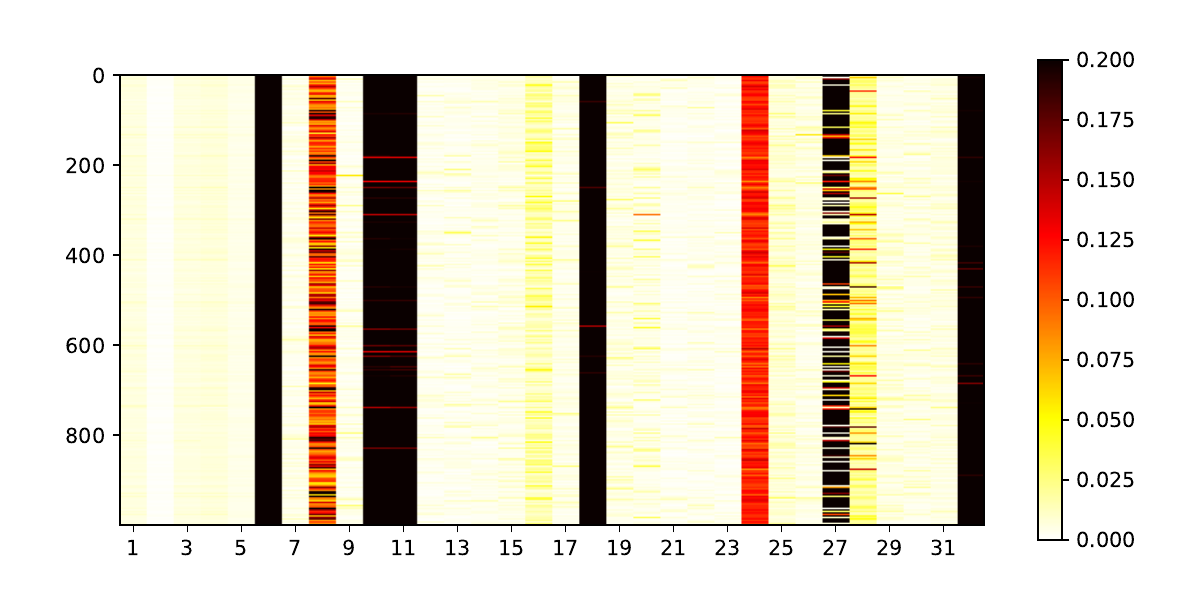}
        \caption{The attention map of $\mathcal{D}^\text{1k}_\text{LLM}$.}
        \label{fig:attention_llmout}
    \end{subfigure}
    \caption{The attention maps of $\mathcal{D}^\text{1k}_\text{clean}$, $\mathcal{D}^{1k}_\text{CLIP}$ and $\mathcal{D}^\text{1k}_\text{LLM}$. Due to space limitations, we select only one layer (the 16th layer of the LLM) and take the maximum value in the multi-head attention. The probe questions are all ``Is there a clock?''. The attention maps of adversarial examples differed significantly from those of the clean examples on certain sensitive tokens (\eg, the 27th and 28 tokens), which are good indicators.}
    \label{fig:attention_diff_clean_adv}
\end{figure}

\begin{algorithm}[t]
    \SetKwInput{KwModel}{Model}
    \caption{The steps of PIP using SVM.}
    \label{alg:example}
    \KwIn{Clean reference dataset $\mathcal{D}^{clean}_{ref}$, $\mathcal{D}_{test}$ to be tested}
    \KwModel{The large vision-language model $f$, where $f_{att}(I, Q)$ outputs the LVLM's attention map}
    \KwData{The irrelevant probe question $Q_p$}
    \KwOut{The predict results of $\mathcal{D}_{test}$, where 0 denotes clean example and 1 denotes adversarial examples}
        $\mathcal{D}^{adv}_{ref}=\{\}$\; 
        \For{image $I_j$ and question $Q_j \in\mathcal{D}^{clean}_{ref}$}
        {
            \tcc{AA denotes a adversarial attack which returns the adversarial image}
            $I'_j=\text{AA}(I_j, Q_j)$\; 
            $\mathcal{D}^{adv}_{ref}=\mathcal{D}^{adv}_{ref}+\{I'_j\}$\;
        }
        \tcc{The train data ($X$) and label ($y$) of SVM}
        $X=\{\}$\;
        $y=\{\}$\; 
        \For{image $I_j\in\{\mathcal{D}^{clean}_{ref}+\mathcal{D}^{adv}_{ref}\}$}
        {
            $X=X+\{f_{att}(I_j, Q_p)\}$\;
            \tcc{Adversarial examples are labeled 1, otherwise 0}
            $y=y+\{\mathbb{I}(I_j\in\mathcal{D}^{adv}_{ref})\}$\;
        }
        \tcc{Train the classifier $h$ with SVM}
        $h=\text{SVM}(X, y)$\;
        \tcc{Using the classifier $h$ to predict}
        $Y_{pred}=\{\}$\;
        \For{$I_j\in\mathcal{D}_{test}$}{
            $Y_{pred}=Y_{pred}+\{h(f_{att}(I_j, Q_p))\}$\;
        }
    \Return{$Y_{pred}$}\;
\end{algorithm}

As previously identified in \cref{sec:yesno_regular}, yes/no type questions exhibit regular attention patterns. This section explores the differences in attention patterns between clean and adversarial examples for yes/no questions.

We randomly selected 1000 images and questions from dataset $\mathcal{D}$ to create dataset $\mathcal{D}^\text{1k}_\text{clean}$. Using PGD on the white-box model, we attack $\mathcal{D}^\text{1k}_\text{clean}$ to generate the adversarial example datasets $\mathcal{D}^{1k}_\text{CLIP}$ and $\mathcal{D}^\text{1k}_\text{LLM}$. For the PGD attack, the number of steps was set to $20$, with a perturbation size per step of $\alpha=2/255$, and a total perturbation limit of $\epsilon_\infty=8/255$. We consider two methods of attack. For $\mathcal{D}^\text{1k}_\text{LLM}$, an untargeted attack was conducted on the LLM’s logit cross-entropy. Obtaining the LLM component of an LVLM can be challenging, whereas accessing its visual encoder (typically CLIP \cite{radford2021learning} or EVA-CLIP \cite{sun2023eva}) is comparatively easier for an adversary. Therefore, for $\mathcal{D}^\text{1k}_\text{CLIP}$, the mean square error (MSE) loss function was employed to conduct an untargeted attack on the output feature of the CLIP or EVA-CLIP visual encoder.

We \textbf{randomly} select a yes/no question as the probe question, for example, ``\textbf{\emph{Is there a clock?}}''. Images from $\mathcal{D}^\text{1k}_\text{clean}$, $\mathcal{D}^{1k}_\text{CLIP}$ and $\mathcal{D}^\text{1k}_\text{LLM}$, along with the probe question, were fed into InstructBLIP Vicuna-7B, with their attention maps displayed in \cref{fig:attention_diff_clean_adv}. Clearly, \Cref{fig:attention_clean} exhibits significant differences when compared to \cref{fig:attention_vout} and \cref{fig:attention_llmout}. Specifically, the 27th token in \cref{fig:attention_clean} is predominantly black, whereas \Cref{fig:attention_llmout} shows a reduced percentage of black, and \cref{fig:attention_vout} contains very little black. Furthermore, the attention for the 28th token in \cref{fig:attention_vout} and \cref{fig:attention_llmout} is more pronounced than in \cref{fig:attention_clean}.

Although \cref{fig:attention_diff_clean_adv} illustrates results from just one layer of the LLM, a significant difference between the clean and adversarial examples is already evident. A similar phenomenon is widely observed across other layers of the LLM. It indicates that the attention patterns of clean and adversarial examples are straightforward and likely to be linearly separable, thus motivating the usage of simple linear classifiers for discrimination as our detector.

\subsection{Distinguishing Attention Maps via Lightweight Support Vector Machine}
\label{sec:svm}

As discussed in \cref{sec:yesno_between_clean_adv}, the use of an irrelevant probe question has been identified as a viable method for distinguishing between clean and adversarial examples. A simple linear classifier, precisely a support vector machine (SVM) \cite{boser1992training}, has been employed to differentiate between clean and adversarial examples.

The steps of our PIP method using SVM are outlined in \cref{alg:example} and \cref{fig:overall}. We conducted an attack using PGD on the cross-entropy loss function of the LLM’s logit output on $\mathcal{D}^{clean}_{ref}$, resulting in the dataset $\mathcal{D}^{adv}_{ref}$. These clean and adversarial images, along with an irrelevant probe question, were fed into the LVLM to extract their attention maps. Utilizing these attention maps, we trained a support vector machine (SVM) $h$ to discern whether the tested examples are adversarial or not.

Acknowledging that adversarial examples are likely less prevalent than clean examples in practical scenarios (due to fewer malicious users), the mixing ratios $M_{clean}:M_{adv}$ were established at $1000:100$ and $1000:1000$. The clean reference dataset $\mathcal{D}_{ref}$ comprises 5000 samples ($N=5000$). The PGD attack encompasses a 20-step iteration, with a step learning rate of $2/255$ and an overall perturbation limit of $\epsilon_\infty=8/255$. 

\begin{table}[t]
  \caption{Results of using PIP with SVM to detect adversarial examples. In this table, $\mathcal{D}_{ref}$ and $\mathcal{D}_{test}$ are from the same COCO dataset.}
  \label{tab:svm_coco}
  \resizebox{\linewidth}{!}{
  \begin{tabular}{c|c|cccc}
    \toprule
    Attack & $M_{clean}/M_{adv}$ & Precision & Recall & Accuracy & F1-score \\
    \midrule
    \multirow{2}{*}{$\mathcal{D}^{CLIP}_{adv}$} & $1000/1000$ & 90.91 & 100.00 & 95.00 & 95.24 \\
    & $1000/100$ & 50.00 & 100.00 & 90.91 & 66.67 \\
    \midrule
    \multirow{2}{*}{$\mathcal{D}^{LLM}_{adv}$} & $1000/1000$ & 98.97 & 96.50 & 97.75 & 97.72 \\
    & $1000/100$ & 90.48 & 95.00 & 98.64 & 92.68 \\
  \bottomrule
\end{tabular}
}
\end{table}

The results of detecting adversarial examples using PIP with SVM are presented in \cref{tab:svm_coco}. For $\mathcal{D}^{LLM}_{adv}$, given that the SVM was trained while also attacking the output of LLM, it demonstrates excellent capability in recognizing adversarial examples, achieving very high performance. For $\mathcal{D}^{CLIP}_{adv}$, despite training the SVM with a different adversarial attack, it remains capable of recognizing adversarial examples. However, $\mathcal{D}^{CLIP}_{adv}$ exhibits a high false alarm rate (10\%) due to the disparity in attack methods used during training SVM and testing examples. With a clean-to-adversarial example ratio of $10:1$, examples with false alarms constitute half of the total alarms, resulting in reduced precision. It is astonishing that our simple PIP achieves impressive results in adversarial example detection.

\subsection{Exploring the PIP's Decision-making Process with Decision Trees}

In previous sections, PIP with SVM was used to detect adversarial samples. However, visualizing the SVM's decision-making process is challenging due to the high-dimensional space of the attention maps. In this section, the decision tree (DT) is used as an intuitive alternative to SVM.

\begin{figure}[t]
    \centering
    \includegraphics[width=\linewidth]{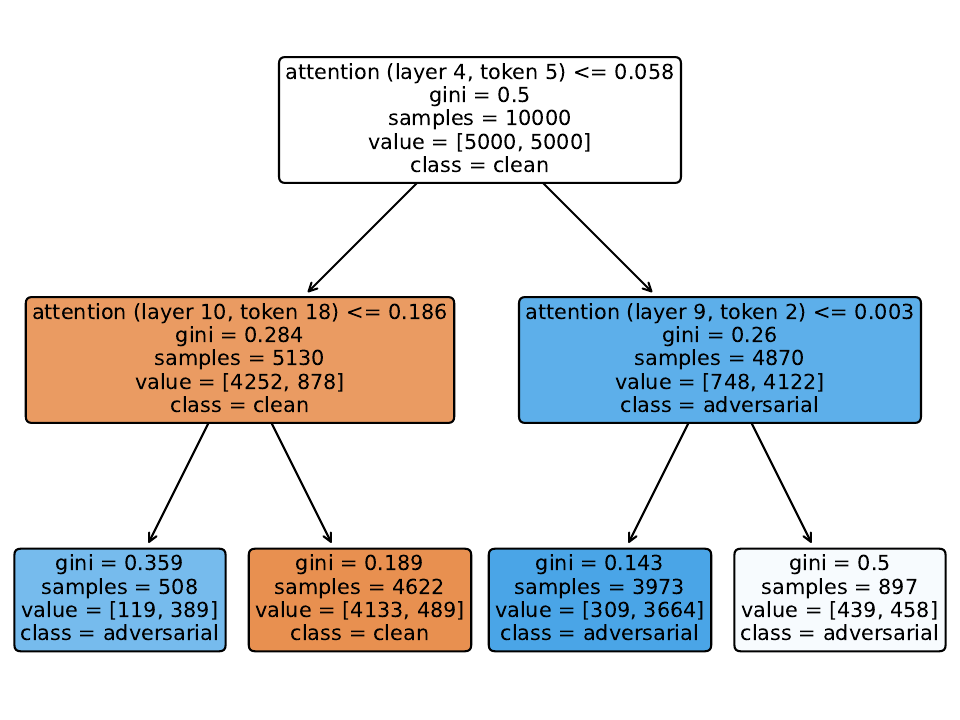}
    \caption{PIP with the decision-making process of decision trees. the DT(depth=2) linearly distinguishes between clean and adversarial examples only by the two feature dimensions of the attention maps. }
    \label{fig:plot_dt}
\end{figure}

Decision tree operates by recursively partitioning a dataset into increasingly smaller subsets and facilitates the generation of easily understandable rules. To visualize the decision-making process, a DT with a depth of 2 is trained. The input for the DT consists of attention maps with 1024 dimensions ($32$ layers $\times$ $32$ tokens, with the maximum value in the multi-head attention dimension). The DT(depth=2) makes a linear decision based on two dimensions, as illustrated in \cref{fig:plot_dt}. Despite its limited depth of 2, the DT successfully detects adversarial examples, as evidenced in \cref{tab:dt_coco}. Increasing the DT's depth could enhance its performance. However, this work only focuses on visualizing the decision-making process using DT(depth=2) to demonstrate PIP's effectiveness.

\section{In-depth Analyses on PIP}

Having initially validated PIP's effectiveness, we then explored its generalizability across various settings and endeavored to understand and analyze our detection method and the intrinsic mechanisms underlying the differences in attention patterns.

We obtained 5,000 $\mathcal{D}^{adv}_{ref}$ images from adversarial attacks on $\mathcal{D}^{clean}_{ref}$. Afterward, an SVM was trained using the attention pattern of 10,000 images on the probe question. In \cref{sec:abalation_test_dataset}, \cref{sec:abalation_attack_method} and \cref{sec:abalation_attack_parameter}, we try the results of this SVM on different test data distributions (training SVM on COCO while test adversarial examples on ImageNet), attack methods (PGD to C\&W), and attack parameters ($\epsilon_\infty$). \Cref{sec:abalation_other_lvlm} explores the results of PIP on other LVLMs (different models require different SVMs to be trained because of different attention patterns), and \Cref{sec:abalation_multipie_svm} explores the use of multiple SVM fusion decisions for the same model. Finally, we explore a real-world black-box attack scenario in \cref{sec:abalation_attack_bard}. It is important to emphasize that for all experiments (except \cref{sec:abalation_multipie_svm}), we trained the SVMs using the \textbf{same data} (PGD $\epsilon_\infty=8/255$ on 5,000 images from COCO $\mathcal{D}_{ref}$ targets the output of LLM) and the \textbf{same probe question} ``Is there a clock?'' to extract attention maps (\Cref{sec:abalation_multipie_svm} used more probe questions than one). Note that using the same unified PIP detector for different settings is extremely challenging.

\subsection{Generalization of Our Adversarial Examples Detection Method across Datasets}
\label{sec:abalation_test_dataset}
In \cref{sec:svm}, a portion of the adversarial examples was successfully detected in the dataset $\mathcal{D}_{test}$. However, there exists a potential issue as both $\mathcal{D}_{ref}$ and $\mathcal{D}_{test}$ derive from $\mathcal{D}$ (\ie, COCO dataset). In practical applications, access to the dataset of user input images may be unavailable, making it crucial to assess the generalization performance when $\mathcal{D}_{ref}$ and $\mathcal{D}_{test}$ do not align.

In this section, $\mathcal{D}_{test}$ is replaced with the ImageNet dataset, while maintaining $\mathcal{D}_{ref}$ as is. Specifically, an image from each of ImageNet’s 1000 classes was randomly selected, and its label was used to generate a corresponding question using the template ``Is there a/an \{label\}?''. This process forms the dataset $\mathcal{D}_{clean}$, and then, following the previous method, these 1000 images and questions were attacked to create the dataset $\mathcal{D}_{adv}$. $\mathcal{D}_{clean}$ and $\mathcal{D}_{adv}$ were mixed according to $M_{clean}:M_{adv}$, with the detection of adversarial examples performed using the method described in \cref{sec:svm}. The results are presented in \cref{tab:svm_imagenet}. Despite the inconsistency between $\mathcal{D}_{ref}$ (from COCO) and $\mathcal{D}_{test}$ (from Imagenet-1K), PIP maintains high accuracy (over 90\%). This indicates that PIP's effectiveness is not contingent on a specific dataset distribution and can be generalized across different datasets. In a practical scenario, $\mathcal{D}_{ref}$ can be generated from a given dataset, with PIP employed to detect adversarial examples in $\mathcal{D}_{test}$,  irrespective of $\mathcal{D}_{test}$'s distribution during the generation of $\mathcal{D}_{ref}$.

\begin{table}[t]
  \caption{Results of using PIP with DT(depth=2) to detect adversarial examples. The purpose of this table is not performance, but to show that PIP does indeed find linearly separable features between clean and adversarial examples, even through DT(depth=2). Its decision-making process is shown in \cref{fig:plot_dt}.}
  \label{tab:dt_coco}
  \resizebox{\linewidth}{!}{
  \begin{tabular}{c|c|cccc}
    \toprule
    Attack & $M_{clean}/M_{adv}$ & Precision & Recall & Accuracy & F1-score \\
    \midrule
    \multirow{2}{*}{$\mathcal{D}^{LLM}_{adv}$} & $1000/1000$ & 85.74 & 90.20 & 87.60 & 87.91 \\
    & $1000/100$ & 37.76 & 91.00 & 85.55 & 53.37 \\
  \bottomrule
\end{tabular}
}
\end{table}

\begin{table}[t]
  \caption{Results of using PIP with SVM to detect adversarial examples. In this table, $\mathcal{D}_{ref}$ and $\mathcal{D}_{test}$ are from COCO and ImageNet-1K, respectively.}
  \label{tab:svm_imagenet}
  \resizebox{\linewidth}{!}{
  \begin{tabular}{c|c|cccc}
    \toprule
    Attack & $M_{clean}/M_{adv}$ & Precision & Recall & Accuracy & F1-score \\
    \midrule
    \multirow{2}{*}{$\mathcal{D}^{CLIP}_{adv}$} & $1000/1000$ & 90.91 & 100.00 & 95.00 & 95.24 \\
    & $1000/100$ & 50.00 & 100.00 & 90.91 & 66.67 \\
    \midrule
    \multirow{2}{*}{$\mathcal{D}^{LLM}_{adv}$} & $1000/1000$ & 90.37 & 93.80 & 91.90 & 92.05 \\
    & $1000/100$ & 48.72 & 95.00 & 90.45 & 64.41 \\
  \bottomrule
\end{tabular}
}
\end{table}

\subsection{Generalizability of PIP over Different Attack Methods}
\label{sec:abalation_attack_method}

\begin{table}[t]
  \caption{Generalizability of PIP over different attack method C\&W. In this table, we only consider attacks the LLM's outputs (\ie, $\mathcal{D}^{LLM}_{adv}$).}
  \label{tab:svm_cw}
  \resizebox{\linewidth}{!}{
  \begin{tabular}{c|c|cccc}
    \toprule
    $\mathcal{D}_{test}$ & $M_{clean}/M_{adv}$ & Precision & Recall & Accuracy & F1-score \\
    \midrule
    \multirow{2}{*}{COCO} & $1000/1000$ & 98.99 & 97.80 & 98.40 & 98.39 \\
    & $1000/100$ & 90.83 & 99.00 & 99.00 & 94.74 \\
    \midrule
    \multirow{2}{*}{ImageNet-1K} & $1000/1000$ & 90.55 & 95.80 & 92.90 & 93.10 \\
    & $1000/100$ & 49.24 & 97.00 & 90.64 & 65.32 \\
  \bottomrule
\end{tabular}
}

\end{table}

In prior experiments, the generalizability of PIP across various attack targets, such as $\mathcal{D}^{CLIP}_{adv}$ and $\mathcal{D}^{LLM}_{adv}$) was established. However, these experiments were exclusively based on the PGD attack method. The ability of SVMs trained on PGD-generated adversarial examples to generalize to unknown attack methods in real-world scenarios remains a critical aspect of PIP's utility.

This section examines the effectiveness of PIP against another popular attack method C\&W beyond PGD. For the C\&W attack, the process iterated over 50 steps with a perturbation of 0.01 per step, equivalent to 2.55/255. \Cref{tab:svm_cw} displays the experimental results. Even against unknown attacks (C\&W) and on datasets distinct from $\mathcal{D}_{ref}$ (ImageNet-1K), PIP demonstrates superior performance.

\subsection{Generalization Performance of PIP on Different Attack Parameters}
\label{sec:abalation_attack_parameter}
In the training process of SVM and prior experiments involving PGD attacks, a consistent perturbation strength of $\epsilon_\infty=8$ was employed. The aim is to ascertain PIP's generalizability across weaker and stronger PGD attacks. \Cref{tab:svm_pgd_eps} examines PIP's efficacy on $\mathcal{D}_{test}$ examples under three perturbation levels $\epsilon_\infty=\{2, 4, 16\}$, revealing that PIP maintains a 40\% recall even with a minimal perturbation of $\epsilon_\infty=2$. With an $\epsilon_\infty=4$ attack, PIP achieves over 80\% recall. Under a significantly strong attack of $\epsilon_\infty=16$, PIP is able to detect nearly all adversarial examples. \Cref{tab:svm_pgd_eps} demonstrates PIP’s generalizability to both weaker and stronger attacks. %, indicating a level of generalizability.

\begin{table}[t]
  \caption{Generalizability of PIP over different methods than PGD attacks. In this table, we only consider attacks the LLM's outputs (\ie, $\mathcal{D}^{LLM}_{adv}$) on COCO.}
  \label{tab:svm_pgd_eps}
  \resizebox{\linewidth}{!}{
  \begin{tabular}{c|c|cccc}
    \toprule
    $\epsilon_\infty$ & $M_{clean}/M_{adv}$ & Precision & Recall & Accuracy & F1-score \\
    \midrule
    \multirow{2}{*}{2/255} & $1000/1000$ & 97.45 & 38.20 & 68.60 & 54.89 \\
    & $1000/100$ & 80.00 & 40.00 & 93.64 & 53.33 \\
    \midrule
    \multirow{2}{*}{4/255} & $1000/1000$ & 98.81 & 82.80 & 90.90 & 90.10 \\
    & $1000/100$ & 89.25 & 83.00 & 97.55 & 86.01 \\
    \midrule
    \multirow{2}{*}{16/255} & $1000/1000$ & 98.99 & 98.50 & 98.75 & 98.75 \\
    & $1000/100$ & 90.83 & 99.00 & 99.00 & 94.74 \\
  \bottomrule
\end{tabular}
}

\end{table}

\subsection{Results of PIP on other LVLMs}
\label{sec:abalation_other_lvlm}
In \cref{tab:svm_coco} and \cref{tab:svm_imagenet}, InstructBLIP Vicuna-7B is utilized as the LVLM. Assessing whether other LVLMs can employ PIP to detect adversarial examples is crucial for evaluating PIP’s generalizability. This section examines other versions of BLIP-2 and InstructBLIP to evaluate PIP's performance on various LVLMs. For each model, PIP initially extracts the attention maps from $\mathcal{D}_{ref}$, followed by SVM training. PIP’s transferability across LVLMs is impossible due to varying attention map configurations and dimensions among LVLMs, such as [32, 32] for Vicuna-7B, [40, 40] for Vicuna-13B, [24, 32] for $\text{FlanT5}_\text{XL}$, [24, 64] for $\text{FlanT5}_\text{XL}$, with the first number representing the LLM’s layers and the second the multi-head attention count. Therefore, a separate PIP detector must be trained on each LVLMs. For BLIP-2 and InstructBLIP, the analysis includes both decoder-only LLMs (\eg, OPT, Vicuna) and encoder-decoder LLMs (\eg, FlanT5). In decoder-only LLMs, we focus on the attention maps of 32 image tokens during the generation of the first word. In encoder-decoder LLMs, we focus on the cross-attention maps of 32 image tokens at the first word’s generation. Within the multi-head attention dimension, only the head with the highest attention is selected to reduce the attention map dimensions. \Cref{tab:svm_other_model} presents the results of applying PIP with SVM across a broader range of LVLMs and demonstrates the effectiveness of PIP on different LVLMs.

\begin{table}[t]
  \caption{Results of PIP on other LVLMs besides InstructBLIP Vicuna-7B. In this table, we only consider attacks the LLM's outputs (\ie, $\mathcal{D}^{LLM}_{adv}$) on COCO.}
  \label{tab:svm_other_model}
  \resizebox{\linewidth}{!}{
  \begin{tabular}{c|c|cccc}
    \toprule
    LVLM & $M_{clean}/M_{adv}$ & Precision & Recall & Accuracy & F1-score \\
    \midrule
    InstructBLIP & $1000/1000$ & 96.19 & 96.00 & 96.10 & 96.10 \\
    ($\text{FlanT5}_\text{XL}$) & $1000/100$ & 71.85 & 97.00 & 96.27 & 82.55 \\
    \midrule
    InstructBLIP & $1000/1000$ & 97.48 & 96.90 & 97.20 & 97.19 \\
    ($\text{FlanT5}_\text{XXL}$) & $1000/100$ & 79.67 & 98.00 & 97.55 & 87.89 \\
    \midrule
    InstructBLIP & $1000/1000$ & 98.79 & 98.20 & 98.50 & 98.50 \\
     (Vicuna-13B) & $1000/100$ & 89.19 & 99.00 & 98.82 & 93.84 \\
    \midrule
    BLIP-2  & $1000/1000$ & 96.36 & 95.20 & 95.80 & 95.77 \\
    (OPT-2.7B) & $1000/100$ & 72.93 & 97.00 & 96.45 & 83.26 \\
    \midrule
    BLIP-2  & $1000/1000$ & 95.78 & 95.40 & 95.60 & 95.59 \\
    (OPT-6.7B) & $1000/100$ & 69.12 & 94.00 & 95.64 & 79.66 \\
    \midrule
    BLIP-2  & $1000/1000$ & 94.73 & 91.70 & 93.30 & 93.19 \\
    ($\text{FlanT5}_\text{XL}$) & $1000/100$ & 64.58 & 93.00 & 94.73 & 76.23 \\
  \bottomrule
\end{tabular}
}
\end{table}

\subsection{Alleviating the High False Alarm Rate Issue}
\label{sec:abalation_multipie_svm}
\begin{table}[t]
  \caption{Results of PIP with multipie SVMs. AR (Alarm Rules) $i/j$ means that among $j$ SVMs of $j$ probe questions, the alarmed SVMs $\ge i$ are determined as adversarial examples.}
  \label{tab:multi_svm}
  \resizebox{\linewidth}{!}{
  \begin{tabular}{c|c|c|c|cccc}
    \toprule
    $\mathcal{D}_{test}$ & AR & Attack & $M_{clean}/M_{adv}$ & Precision & Recall & Accuracy & F1-score \\
    \midrule
    \multirow{4}{*}{COCO} & \multirow{4}{*}{$2/3$} & \multirow{2}{*}{$\mathcal{D}^{CLIP}_{adv}$} & $1000/1000$ & 98.81 & 100.00 & 99.40 & 99.40 \\
    & & & $1000/100$ & 89.29 & 100.00 & 98.91 & 94.34 \\
    \cmidrule{3-8}
    & & \multirow{2}{*}{$\mathcal{D}^{LLM}_{adv}$} & $1000/1000$ & 98.78 & 96.80 & 97.80 & 97.78 \\
    & & & $1000/100$ & 88.99 & 97.00 & 98.64 & 92.82 \\
    \midrule
    \multirow{4}{*}{IN-1K} & \multirow{4}{*}{$3/3$} & \multirow{2}{*}{$\mathcal{D}^{CLIP}_{adv}$} & $1000/1000$ & 94.61 & 100.00 & 97.15 & 97.23 \\
    & & & $1000/100$ & 63.69 & 100.00 & 94.82 & 77.82 \\
    \cmidrule{3-8}
    & & \multirow{2}{*}{$\mathcal{D}^{LLM}_{adv}$} & $1000/1000$ & 94.06 & 90.30 & 92.30 & 92.14 \\
    & & & $1000/100$ & 61.74 & 92.00 & 94.09 & 73.90 \\
  \bottomrule
\end{tabular}
}

\end{table}

The primary advantage of PIP is its high recall rate, which is critical for LVLMs. Adversaries can manipulate the model using adversarial examples, potentially leading to significant public opinion risks if the model generates content that violates morality or law. However, as indicated in \cref{tab:svm_coco}, PIP experiences a high false alarm rate (about 10\%) when the attack method used to train the SVM differs from that used on the test examples. This false alarm rate is particularly high when the ratio of clean to adversarial examples is $1000:100$, resulting in half of the alarms being false. Such a high false alarm rate could jeopardize PIP's practical application, as normal user images may be incorrectly flagged, degrading the user experience.

Our fundamental approach involves conducting a focused review of samples triggering alerts. A straightforward method is to apply three instead of one probe question to these targeted samples. In scenarios with low false alarm rates, like in \cref{tab:svm_coco}, an image is classified as an adversarial example if it triggers at least two alarms out of three probe questions. In cases with high false alarm rates, such as in \cref{tab:svm_imagenet}, an image is deemed an adversarial example only if alarms are triggered on all three probe questions. Besides the existing probe question ``Is there a clock?'', two additional questions, ``Is this in the United States?'' and ``Is this photo an action shot?'' were randomly selected, and two more SVMs were trained accordingly.

The outcomes are detailed in \cref{tab:multi_svm}, which, compared to \cref{tab:svm_coco} and \cref{tab:svm_imagenet}, exhibits a notable reduction in the false alarm rate, as indicated by a substantial increase in precision. Moreover, \Cref{tab:multi_svm} maintains an exceptionally high recall rate, thus reducing the false alarm rate while preserving the detection of adversarial examples.

Employing additional SVMs will increase computational demands. For a user-input image, one SVM judgment necessitates only one extra inference beyond the user-input question to derive the attention map, while three SVM judgments necessitate three extra inferences. Fortunately, not all images require multiple inferences. It suffices to infer all images once and selectively focus on certain suspicious images. %, which may necessitate a second or even third inference.

\subsection{Generalization Performance of PIP on Actual Black-box Attacks}
\label{sec:abalation_attack_bard}
In earlier experiments, adversarial examples were generated through the white-box attacks. However, in real-world scenarios, obtaining the model's weights and executing white-box attacks is challenging for users. Consequently, numerous studies have investigated black-box attacks. Exploring PIP's effectiveness against black-box attack-generated adversarial examples is worthwhile.

Attack-Bard \cite{dong2023robust} used black-box attacks on the NIPS2017 dataset to generate adversarial examples, successfully compromising major commercial models like ChatGPT-4V, Google’s Bard, Bing Chat, and ERNIE Bot (with approximately 45\% attack success on ChatGPT-4V). This section involves selecting 200 original images from the NIPS2017 dataset along with corresponding adversarial images generated by Attack-Bard to create the dataset $\mathcal{D}_{test}$, using PIP to detect adversarial examples on InstructBLIP.

This constitutes a comprehensive evaluation of PIP, as in this experimental setup, the constant factor is the model used in both training and detection phases (necessary due to varying attention patterns across models). Beyond this, PIP remains uninformed about other aspects like the distribution of user input data, attack methods, parameters, targets, and the models used for the attack. \Cref{tab:pip_attack_bard} affirms PIP's generalizability in authentic black-box attack scenarios. In the context of black-box attacks, PIP maintains a recall rate exceeding 95\% and a precision greater than 80\%.

\begin{table}[t]
  \caption{Results of PIP on detecting adversarial examples generated by black-box Attack-Bard on NIPS2017 dataset. The $M_{clean}/M_{adv}$ is $200:200$.}
  \label{tab:pip_attack_bard}
  \resizebox{\linewidth}{!}{
  \begin{tabular}{c|ccccc}
    \toprule
    LVLM & Precision & Recall & Accuracy & F1-score \\
    \midrule
    InstructBLIP $\text{FlanT5}_\text{XL}$ & 90.09 & 100.00 & 94.50 & 94.79 \\
    \midrule
    InstructBLIP $\text{FlanT5}_\text{XXL}$ & 90.32 & 98.00 & 93.75 & 94.00 \\
    \midrule
    InstructBLIP Vicuna-7B & 94.79 & 100.00 & 97.25 & 97.32 \\
    \midrule
    InstructBLIP Vicuna-13B & 96.10 & 98.50 & 97.25 & 97.28 \\
  \bottomrule
\end{tabular}
}
\end{table}

\section{Conclusion}
In this paper, we introduce PIP, a new and simple method for detecting adversarial examples in LVLMs. Although PIP is simple and whimsical, it achieves impressive results on recent, popular LVLMs like BLIP-2 and InstructBLIP, achieving high recall rates of adversarial examples with low false alarms among clean examples. For detected adversarial examples, post-processing measures such as focusing on alert examples, denying answers, and implementing adversarial defenses can enhance the security and robustness of LVLMs, thereby mitigating public and legal risks.

\begin{acks}
This work was supported by the National Natural Science Foundation of China (62376024, 62171313), the Young Elite Scientists Sponsorship Program by CAST (2023QNRC001) and Beijing National Research Center for Information Science and Technology (BNRist).
\end{acks}

\bibliographystyle{ACM-Reference-Format}
\balance
\bibliography{sample-base}


\begin{thebibliography}{35}

%%% the Web address.

\ifx \showCODEN    \undefined \def \showCODEN     #1{\unskip}     \fi
\ifx \showDOI      \undefined \def \showDOI       #1{#1}\fi
\ifx \showISBNx    \undefined \def \showISBNx     #1{\unskip}     \fi
\ifx \showISBNxiii \undefined \def \showISBNxiii  #1{\unskip}     \fi
\ifx \showISSN     \undefined \def \showISSN      #1{\unskip}     \fi
\ifx \showLCCN     \undefined \def \showLCCN      #1{\unskip}     \fi
\ifx \shownote     \undefined \def \shownote      #1{#1}          \fi
\ifx \showarticletitle \undefined \def \showarticletitle #1{#1}   \fi
\ifx \showURL      \undefined \def \showURL       {\relax}        \fi
% invisible to TeX
\providecommand\bibfield[2]{#2}
\providecommand\bibinfo[2]{#2}
\providecommand\natexlab[1]{#1}
\providecommand\showeprint[2][]{arXiv:#2}

\bibitem[Alayrac et~al\mbox{.}(2022)]%
        {alayrac2022flamingo}
\bibfield{author}{\bibinfo{person}{Jean-Baptiste Alayrac}, \bibinfo{person}{Jeff Donahue}, \bibinfo{person}{Pauline Luc}, \bibinfo{person}{Antoine Miech}, \bibinfo{person}{Iain Barr}, \bibinfo{person}{Yana Hasson}, \bibinfo{person}{Karel Lenc}, \bibinfo{person}{Arthur Mensch}, \bibinfo{person}{Katherine Millican}, \bibinfo{person}{Malcolm Reynolds}, {et~al\mbox{.}}} \bibinfo{year}{2022}\natexlab{}.
\newblock \showarticletitle{Flamingo: a visual language model for few-shot learning}.
\newblock \bibinfo{journal}{\emph{Advances in neural information processing systems}}  \bibinfo{volume}{35} (\bibinfo{year}{2022}), \bibinfo{pages}{23716--23736}.
\newblock


\bibitem[Bailey et~al\mbox{.}(2023)]%
        {bailey2023image}
\bibfield{author}{\bibinfo{person}{Luke Bailey}, \bibinfo{person}{Euan Ong}, \bibinfo{person}{Stuart Russell}, {and} \bibinfo{person}{Scott Emmons}.} \bibinfo{year}{2023}\natexlab{}.
\newblock \showarticletitle{Image hijacks: Adversarial images can control generative models at runtime}.
\newblock \bibinfo{journal}{\emph{arXiv preprint arXiv:2309.00236}} (\bibinfo{year}{2023}).
\newblock


\bibitem[Boser et~al\mbox{.}(1992)]%
        {boser1992training}
\bibfield{author}{\bibinfo{person}{Bernhard~E Boser}, \bibinfo{person}{Isabelle~M Guyon}, {and} \bibinfo{person}{Vladimir~N Vapnik}.} \bibinfo{year}{1992}\natexlab{}.
\newblock \showarticletitle{A training algorithm for optimal margin classifiers}. In \bibinfo{booktitle}{\emph{Proceedings of the fifth annual workshop on Computational learning theory}}. \bibinfo{pages}{144--152}.
\newblock


\bibitem[Carlini et~al\mbox{.}(2024)]%
        {carlini2024aligned}
\bibfield{author}{\bibinfo{person}{Nicholas Carlini}, \bibinfo{person}{Milad Nasr}, \bibinfo{person}{Christopher~A Choquette-Choo}, \bibinfo{person}{Matthew Jagielski}, \bibinfo{person}{Irena Gao}, \bibinfo{person}{Pang Wei~W Koh}, \bibinfo{person}{Daphne Ippolito}, \bibinfo{person}{Florian Tramer}, {and} \bibinfo{person}{Ludwig Schmidt}.} \bibinfo{year}{2024}\natexlab{}.
\newblock \showarticletitle{Are aligned neural networks adversarially aligned?}
\newblock \bibinfo{journal}{\emph{Advances in Neural Information Processing Systems}}  \bibinfo{volume}{36} (\bibinfo{year}{2024}).
\newblock


\bibitem[Carlini and Wagner(2017)]%
        {carlini2017towards}
\bibfield{author}{\bibinfo{person}{Nicholas Carlini} {and} \bibinfo{person}{David Wagner}.} \bibinfo{year}{2017}\natexlab{}.
\newblock \showarticletitle{Towards evaluating the robustness of neural networks}. In \bibinfo{booktitle}{\emph{2017 ieee symposium on security and privacy (sp)}}. Ieee, \bibinfo{pages}{39--57}.
\newblock


\bibitem[Dai et~al\mbox{.}(2024)]%
        {dai2024instructblip}
\bibfield{author}{\bibinfo{person}{Wenliang Dai}, \bibinfo{person}{Junnan Li}, \bibinfo{person}{Dongxu Li}, \bibinfo{person}{Anthony Meng~Huat Tiong}, \bibinfo{person}{Junqi Zhao}, \bibinfo{person}{Weisheng Wang}, \bibinfo{person}{Boyang Li}, \bibinfo{person}{Pascale~N Fung}, {and} \bibinfo{person}{Steven Hoi}.} \bibinfo{year}{2024}\natexlab{}.
\newblock \showarticletitle{Instructblip: Towards general-purpose vision-language models with instruction tuning}.
\newblock \bibinfo{journal}{\emph{Advances in Neural Information Processing Systems}}  \bibinfo{volume}{36} (\bibinfo{year}{2024}).
\newblock


\bibitem[Dathathri et~al\mbox{.}(2018)]%
        {dathathri2018detecting}
\bibfield{author}{\bibinfo{person}{Sumanth Dathathri}, \bibinfo{person}{Stephan Zheng}, \bibinfo{person}{Tianwei Yin}, \bibinfo{person}{Richard~M Murray}, {and} \bibinfo{person}{Yisong Yue}.} \bibinfo{year}{2018}\natexlab{}.
\newblock \showarticletitle{Detecting adversarial examples via neural fingerprinting}.
\newblock \bibinfo{journal}{\emph{arXiv preprint arXiv:1803.03870}} (\bibinfo{year}{2018}).
\newblock


\bibitem[Dong et~al\mbox{.}(2023)]%
        {dong2023robust}
\bibfield{author}{\bibinfo{person}{Yinpeng Dong}, \bibinfo{person}{Huanran Chen}, \bibinfo{person}{Jiawei Chen}, \bibinfo{person}{Zhengwei Fang}, \bibinfo{person}{Xiao Yang}, \bibinfo{person}{Yichi Zhang}, \bibinfo{person}{Yu Tian}, \bibinfo{person}{Hang Su}, {and} \bibinfo{person}{Jun Zhu}.} \bibinfo{year}{2023}\natexlab{}.
\newblock \showarticletitle{How Robust is Google's Bard to Adversarial Image Attacks?}
\newblock \bibinfo{journal}{\emph{arXiv preprint arXiv:2309.11751}} (\bibinfo{year}{2023}).
\newblock


\bibitem[Gao et~al\mbox{.}(2021)]%
        {gao2021maximum}
\bibfield{author}{\bibinfo{person}{Ruize Gao}, \bibinfo{person}{Feng Liu}, \bibinfo{person}{Jingfeng Zhang}, \bibinfo{person}{Bo Han}, \bibinfo{person}{Tongliang Liu}, \bibinfo{person}{Gang Niu}, {and} \bibinfo{person}{Masashi Sugiyama}.} \bibinfo{year}{2021}\natexlab{}.
\newblock \showarticletitle{Maximum mean discrepancy test is aware of adversarial attacks}. In \bibinfo{booktitle}{\emph{International Conference on Machine Learning}}. PMLR, \bibinfo{pages}{3564--3575}.
\newblock


\bibitem[Goodfellow et~al\mbox{.}(2014)]%
        {goodfellow2014explaining}
\bibfield{author}{\bibinfo{person}{Ian~J Goodfellow}, \bibinfo{person}{Jonathon Shlens}, {and} \bibinfo{person}{Christian Szegedy}.} \bibinfo{year}{2014}\natexlab{}.
\newblock \showarticletitle{Explaining and harnessing adversarial examples}.
\newblock \bibinfo{journal}{\emph{arXiv preprint arXiv:1412.6572}} (\bibinfo{year}{2014}).
\newblock


\bibitem[Huang et~al\mbox{.}(2023)]%
        {10.1145/3581783.3611914}
\bibfield{author}{\bibinfo{person}{Yihao Huang}, \bibinfo{person}{Liangru Sun}, \bibinfo{person}{Qing Guo}, \bibinfo{person}{Felix Juefei-Xu}, \bibinfo{person}{Jiayi Zhu}, \bibinfo{person}{Jincao Feng}, \bibinfo{person}{Yang Liu}, {and} \bibinfo{person}{Geguang Pu}.} \bibinfo{year}{2023}\natexlab{}.
\newblock \showarticletitle{ALA: Naturalness-aware Adversarial Lightness Attack}. In \bibinfo{booktitle}{\emph{Proceedings of the 31st ACM International Conference on Multimedia}} (Ottawa ON, Canada) \emph{(\bibinfo{series}{MM '23})}. \bibinfo{publisher}{Association for Computing Machinery}, \bibinfo{address}{New York, NY, USA}, \bibinfo{pages}{2418–2426}.
\newblock
\showISBNx{9798400701085}
\urldef\tempurl%
\url{https://doi.org/10.1145/3581783.3611914}
\showDOI{\tempurl}


\bibitem[Jin et~al\mbox{.}(2019)]%
        {jin2019bert}
\bibfield{author}{\bibinfo{person}{Di Jin}, \bibinfo{person}{Zhijing Jin}, \bibinfo{person}{Joey~Tianyi Zhou}, {and} \bibinfo{person}{Peter Szolovits}.} \bibinfo{year}{2019}\natexlab{}.
\newblock \showarticletitle{Is bert really robust? natural language attack on text classification and entailment}.
\newblock \bibinfo{journal}{\emph{arXiv preprint arXiv:1907.11932}}  \bibinfo{volume}{2} (\bibinfo{year}{2019}), \bibinfo{pages}{10}.
\newblock


\bibitem[Li et~al\mbox{.}(2023)]%
        {li2023blip}
\bibfield{author}{\bibinfo{person}{Junnan Li}, \bibinfo{person}{Dongxu Li}, \bibinfo{person}{Silvio Savarese}, {and} \bibinfo{person}{Steven Hoi}.} \bibinfo{year}{2023}\natexlab{}.
\newblock \showarticletitle{Blip-2: Bootstrapping language-image pre-training with frozen image encoders and large language models}. In \bibinfo{booktitle}{\emph{International conference on machine learning}}. PMLR, \bibinfo{pages}{19730--19742}.
\newblock


\bibitem[Li et~al\mbox{.}(2020)]%
        {li2020bert}
\bibfield{author}{\bibinfo{person}{Linyang Li}, \bibinfo{person}{Ruotian Ma}, \bibinfo{person}{Qipeng Guo}, \bibinfo{person}{Xiangyang Xue}, {and} \bibinfo{person}{Xipeng Qiu}.} \bibinfo{year}{2020}\natexlab{}.
\newblock \showarticletitle{Bert-attack: Adversarial attack against bert using bert}.
\newblock \bibinfo{journal}{\emph{arXiv preprint arXiv:2004.09984}} (\bibinfo{year}{2020}).
\newblock


\bibitem[Li and Li(2017)]%
        {li2017adversarial}
\bibfield{author}{\bibinfo{person}{Xin Li} {and} \bibinfo{person}{Fuxin Li}.} \bibinfo{year}{2017}\natexlab{}.
\newblock \showarticletitle{Adversarial examples detection in deep networks with convolutional filter statistics}. In \bibinfo{booktitle}{\emph{Proceedings of the IEEE international conference on computer vision}}. \bibinfo{pages}{5764--5772}.
\newblock


\bibitem[Liu et~al\mbox{.}(2024)]%
        {liu2024visual}
\bibfield{author}{\bibinfo{person}{Haotian Liu}, \bibinfo{person}{Chunyuan Li}, \bibinfo{person}{Qingyang Wu}, {and} \bibinfo{person}{Yong~Jae Lee}.} \bibinfo{year}{2024}\natexlab{}.
\newblock \showarticletitle{Visual instruction tuning}.
\newblock \bibinfo{journal}{\emph{Advances in neural information processing systems}}  \bibinfo{volume}{36} (\bibinfo{year}{2024}).
\newblock


\bibitem[Lu et~al\mbox{.}(2023)]%
        {lu2023set}
\bibfield{author}{\bibinfo{person}{Dong Lu}, \bibinfo{person}{Zhiqiang Wang}, \bibinfo{person}{Teng Wang}, \bibinfo{person}{Weili Guan}, \bibinfo{person}{Hongchang Gao}, {and} \bibinfo{person}{Feng Zheng}.} \bibinfo{year}{2023}\natexlab{}.
\newblock \showarticletitle{Set-level guidance attack: Boosting adversarial transferability of vision-language pre-training models}. In \bibinfo{booktitle}{\emph{Proceedings of the IEEE/CVF International Conference on Computer Vision}}. \bibinfo{pages}{102--111}.
\newblock


\bibitem[Luo et~al\mbox{.}(2023)]%
        {luo2023image}
\bibfield{author}{\bibinfo{person}{Haochen Luo}, \bibinfo{person}{Jindong Gu}, \bibinfo{person}{Fengyuan Liu}, {and} \bibinfo{person}{Philip Torr}.} \bibinfo{year}{2023}\natexlab{}.
\newblock \showarticletitle{An image is worth 1000 lies: Transferability of adversarial images across prompts on vision-language models}. In \bibinfo{booktitle}{\emph{The Twelfth International Conference on Learning Representations}}.
\newblock


\bibitem[Ma et~al\mbox{.}(2019)]%
        {ma2019metaadvdet}
\bibfield{author}{\bibinfo{person}{Chen Ma}, \bibinfo{person}{Chenxu Zhao}, \bibinfo{person}{Hailin Shi}, \bibinfo{person}{Li Chen}, \bibinfo{person}{Junhai Yong}, {and} \bibinfo{person}{Dan Zeng}.} \bibinfo{year}{2019}\natexlab{}.
\newblock \showarticletitle{Metaadvdet: Towards robust detection of evolving adversarial attacks}. In \bibinfo{booktitle}{\emph{Proceedings of the 27th ACM International Conference on Multimedia}}. \bibinfo{pages}{692--701}.
\newblock


\bibitem[Madry et~al\mbox{.}(2017)]%
        {madry2017towards}
\bibfield{author}{\bibinfo{person}{Aleksander Madry}, \bibinfo{person}{Aleksandar Makelov}, \bibinfo{person}{Ludwig Schmidt}, \bibinfo{person}{Dimitris Tsipras}, {and} \bibinfo{person}{Adrian Vladu}.} \bibinfo{year}{2017}\natexlab{}.
\newblock \showarticletitle{Towards deep learning models resistant to adversarial attacks}.
\newblock \bibinfo{journal}{\emph{arXiv preprint arXiv:1706.06083}} (\bibinfo{year}{2017}).
\newblock


\bibitem[Meng and Chen(2017)]%
        {meng2017magnet}
\bibfield{author}{\bibinfo{person}{Dongyu Meng} {and} \bibinfo{person}{Hao Chen}.} \bibinfo{year}{2017}\natexlab{}.
\newblock \showarticletitle{Magnet: a two-pronged defense against adversarial examples}. In \bibinfo{booktitle}{\emph{Proceedings of the 2017 ACM SIGSAC conference on computer and communications security}}. \bibinfo{pages}{135--147}.
\newblock


\bibitem[Pang et~al\mbox{.}(2018)]%
        {pang2018towards}
\bibfield{author}{\bibinfo{person}{Tianyu Pang}, \bibinfo{person}{Chao Du}, \bibinfo{person}{Yinpeng Dong}, {and} \bibinfo{person}{Jun Zhu}.} \bibinfo{year}{2018}\natexlab{}.
\newblock \showarticletitle{Towards robust detection of adversarial examples}.
\newblock \bibinfo{journal}{\emph{Advances in neural information processing systems}}  \bibinfo{volume}{31} (\bibinfo{year}{2018}).
\newblock


\bibitem[Papernot et~al\mbox{.}(2016)]%
        {papernot2016limitations}
\bibfield{author}{\bibinfo{person}{Nicolas Papernot}, \bibinfo{person}{Patrick McDaniel}, \bibinfo{person}{Somesh Jha}, \bibinfo{person}{Matt Fredrikson}, \bibinfo{person}{Z~Berkay Celik}, {and} \bibinfo{person}{Ananthram Swami}.} \bibinfo{year}{2016}\natexlab{}.
\newblock \showarticletitle{The limitations of deep learning in adversarial settings}. In \bibinfo{booktitle}{\emph{2016 IEEE European symposium on security and privacy (EuroS\&P)}}. IEEE, \bibinfo{pages}{372--387}.
\newblock


\bibitem[Qi et~al\mbox{.}(2024)]%
        {qi2024visual}
\bibfield{author}{\bibinfo{person}{Xiangyu Qi}, \bibinfo{person}{Kaixuan Huang}, \bibinfo{person}{Ashwinee Panda}, \bibinfo{person}{Peter Henderson}, \bibinfo{person}{Mengdi Wang}, {and} \bibinfo{person}{Prateek Mittal}.} \bibinfo{year}{2024}\natexlab{}.
\newblock \showarticletitle{Visual adversarial examples jailbreak aligned large language models}. In \bibinfo{booktitle}{\emph{Proceedings of the AAAI Conference on Artificial Intelligence}}, Vol.~\bibinfo{volume}{38}. \bibinfo{pages}{21527--21536}.
\newblock


\bibitem[Radford et~al\mbox{.}(2021)]%
        {radford2021learning}
\bibfield{author}{\bibinfo{person}{Alec Radford}, \bibinfo{person}{Jong~Wook Kim}, \bibinfo{person}{Chris Hallacy}, \bibinfo{person}{Aditya Ramesh}, \bibinfo{person}{Gabriel Goh}, \bibinfo{person}{Sandhini Agarwal}, \bibinfo{person}{Girish Sastry}, \bibinfo{person}{Amanda Askell}, \bibinfo{person}{Pamela Mishkin}, \bibinfo{person}{Jack Clark}, {et~al\mbox{.}}} \bibinfo{year}{2021}\natexlab{}.
\newblock \showarticletitle{Learning transferable visual models from natural language supervision}. In \bibinfo{booktitle}{\emph{International conference on machine learning}}. PMLR, \bibinfo{pages}{8748--8763}.
\newblock


\bibitem[Shang et~al\mbox{.}(2023)]%
        {10.1145/3581783.3612337}
\bibfield{author}{\bibinfo{person}{Yu Shang}, \bibinfo{person}{Chen Gao}, \bibinfo{person}{Jiansheng Chen}, \bibinfo{person}{Depeng Jin}, \bibinfo{person}{Huimin Ma}, {and} \bibinfo{person}{Yong Li}.} \bibinfo{year}{2023}\natexlab{}.
\newblock \showarticletitle{Enhancing Adversarial Robustness of Multi-modal Recommendation via Modality Balancing}. In \bibinfo{booktitle}{\emph{Proceedings of the 31st ACM International Conference on Multimedia}} (Ottawa ON, Canada) \emph{(\bibinfo{series}{MM '23})}. \bibinfo{publisher}{Association for Computing Machinery}, \bibinfo{address}{New York, NY, USA}, \bibinfo{pages}{6274–6282}.
\newblock
\showISBNx{9798400701085}
\urldef\tempurl%
\url{https://doi.org/10.1145/3581783.3612337}
\showDOI{\tempurl}


\bibitem[Sun et~al\mbox{.}(2023)]%
        {sun2023eva}
\bibfield{author}{\bibinfo{person}{Quan Sun}, \bibinfo{person}{Yuxin Fang}, \bibinfo{person}{Ledell Wu}, \bibinfo{person}{Xinlong Wang}, {and} \bibinfo{person}{Yue Cao}.} \bibinfo{year}{2023}\natexlab{}.
\newblock \showarticletitle{Eva-clip: Improved training techniques for clip at scale}.
\newblock \bibinfo{journal}{\emph{arXiv preprint arXiv:2303.15389}} (\bibinfo{year}{2023}).
\newblock


\bibitem[Wang et~al\mbox{.}(2024)]%
        {wang2024visionllm}
\bibfield{author}{\bibinfo{person}{Wenhai Wang}, \bibinfo{person}{Zhe Chen}, \bibinfo{person}{Xiaokang Chen}, \bibinfo{person}{Jiannan Wu}, \bibinfo{person}{Xizhou Zhu}, \bibinfo{person}{Gang Zeng}, \bibinfo{person}{Ping Luo}, \bibinfo{person}{Tong Lu}, \bibinfo{person}{Jie Zhou}, \bibinfo{person}{Yu Qiao}, {et~al\mbox{.}}} \bibinfo{year}{2024}\natexlab{}.
\newblock \showarticletitle{Visionllm: Large language model is also an open-ended decoder for vision-centric tasks}.
\newblock \bibinfo{journal}{\emph{Advances in Neural Information Processing Systems}}  \bibinfo{volume}{36} (\bibinfo{year}{2024}).
\newblock


\bibitem[Wang et~al\mbox{.}(2022)]%
        {10.1145/3503161.3547989}
\bibfield{author}{\bibinfo{person}{Yuxuan Wang}, \bibinfo{person}{Jiakai Wang}, \bibinfo{person}{Zixin Yin}, \bibinfo{person}{Ruihao Gong}, \bibinfo{person}{Jingyi Wang}, \bibinfo{person}{Aishan Liu}, {and} \bibinfo{person}{Xianglong Liu}.} \bibinfo{year}{2022}\natexlab{}.
\newblock \showarticletitle{Generating Transferable Adversarial Examples against Vision Transformers}. In \bibinfo{booktitle}{\emph{Proceedings of the 30th ACM International Conference on Multimedia}} (Lisboa, Portugal) \emph{(\bibinfo{series}{MM '22})}. \bibinfo{publisher}{Association for Computing Machinery}, \bibinfo{address}{New York, NY, USA}, \bibinfo{pages}{5181–5190}.
\newblock
\showISBNx{9781450392037}
\urldef\tempurl%
\url{https://doi.org/10.1145/3503161.3547989}
\showDOI{\tempurl}


\bibitem[Xu et~al\mbox{.}(2017)]%
        {xu2017feature}
\bibfield{author}{\bibinfo{person}{Weilin Xu}, \bibinfo{person}{David Evans}, {and} \bibinfo{person}{Yanjun Qi}.} \bibinfo{year}{2017}\natexlab{}.
\newblock \showarticletitle{Feature squeezing: Detecting adversarial examples in deep neural networks}.
\newblock \bibinfo{journal}{\emph{arXiv preprint arXiv:1704.01155}} (\bibinfo{year}{2017}).
\newblock


\bibitem[Yin et~al\mbox{.}(2023)]%
        {yin2023vlattack}
\bibfield{author}{\bibinfo{person}{Ziyi Yin}, \bibinfo{person}{Muchao Ye}, \bibinfo{person}{Tianrong Zhang}, \bibinfo{person}{Tianyu Du}, \bibinfo{person}{Jinguo Zhu}, \bibinfo{person}{Han Liu}, \bibinfo{person}{Jinghui Chen}, \bibinfo{person}{Ting Wang}, {and} \bibinfo{person}{Fenglong Ma}.} \bibinfo{year}{2023}\natexlab{}.
\newblock \showarticletitle{Vlattack: Multimodal adversarial attacks on vision-language tasks via pre-trained models}.
\newblock \bibinfo{journal}{\emph{arXiv preprint arXiv:2310.04655}} (\bibinfo{year}{2023}).
\newblock


\bibitem[Zhang et~al\mbox{.}(2022)]%
        {10.1145/3503161.3547801}
\bibfield{author}{\bibinfo{person}{Jiaming Zhang}, \bibinfo{person}{Qi Yi}, {and} \bibinfo{person}{Jitao Sang}.} \bibinfo{year}{2022}\natexlab{}.
\newblock \showarticletitle{Towards Adversarial Attack on Vision-Language Pre-training Models}. In \bibinfo{booktitle}{\emph{Proceedings of the 30th ACM International Conference on Multimedia}} (Lisboa, Portugal) \emph{(\bibinfo{series}{MM '22})}. \bibinfo{publisher}{Association for Computing Machinery}, \bibinfo{address}{New York, NY, USA}, \bibinfo{pages}{5005–5013}.
\newblock
\showISBNx{9781450392037}
\urldef\tempurl%
\url{https://doi.org/10.1145/3503161.3547801}
\showDOI{\tempurl}


\bibitem[Zhang et~al\mbox{.}(2024)]%
        {10.1145/3626772.3657781}
\bibfield{author}{\bibinfo{person}{Peng-Fei Zhang}, \bibinfo{person}{Zi Huang}, {and} \bibinfo{person}{Guangdong Bai}.} \bibinfo{year}{2024}\natexlab{}.
\newblock \showarticletitle{Universal Adversarial Perturbations for Vision-Language Pre-trained Models}. In \bibinfo{booktitle}{\emph{Proceedings of the 47th International ACM SIGIR Conference on Research and Development in Information Retrieval}} (Washington DC, USA) \emph{(\bibinfo{series}{SIGIR '24})}. \bibinfo{publisher}{Association for Computing Machinery}, \bibinfo{address}{New York, NY, USA}, \bibinfo{pages}{862–871}.
\newblock
\showISBNx{9798400704314}
\urldef\tempurl%
\url{https://doi.org/10.1145/3626772.3657781}
\showDOI{\tempurl}


\bibitem[Zhao et~al\mbox{.}(2024)]%
        {zhao2024evaluating}
\bibfield{author}{\bibinfo{person}{Yunqing Zhao}, \bibinfo{person}{Tianyu Pang}, \bibinfo{person}{Chao Du}, \bibinfo{person}{Xiao Yang}, \bibinfo{person}{Chongxuan Li}, \bibinfo{person}{Ngai-Man~Man Cheung}, {and} \bibinfo{person}{Min Lin}.} \bibinfo{year}{2024}\natexlab{}.
\newblock \showarticletitle{On evaluating adversarial robustness of large vision-language models}.
\newblock \bibinfo{journal}{\emph{Advances in Neural Information Processing Systems}}  \bibinfo{volume}{36} (\bibinfo{year}{2024}).
\newblock


\bibitem[Zhu et~al\mbox{.}(2023)]%
        {zhu2023minigpt}
\bibfield{author}{\bibinfo{person}{Deyao Zhu}, \bibinfo{person}{Jun Chen}, \bibinfo{person}{Xiaoqian Shen}, \bibinfo{person}{Xiang Li}, {and} \bibinfo{person}{Mohamed Elhoseiny}.} \bibinfo{year}{2023}\natexlab{}.
\newblock \showarticletitle{Minigpt-4: Enhancing vision-language understanding with advanced large language models}.
\newblock \bibinfo{journal}{\emph{arXiv preprint arXiv:2304.10592}} (\bibinfo{year}{2023}).
\newblock


\end{thebibliography}

\end{document}